\documentclass[sigconf]{acmart}
\AtBeginDocument{%
  }

\copyrightyear{2026}
\acmYear{2026}
\setcopyright{cc}
\setcctype{by-nc-nd}
\acmConference[KDD '26]{Proceedings of the 32nd ACM SIGKDD Conference on Knowledge Discovery and Data Mining V.1}{August 09--13, 2026}{Jeju Island, Republic of Korea}
\acmBooktitle{Proceedings of the 32nd ACM SIGKDD Conference on Knowledge Discovery and Data Mining V.1 (KDD '26), August 09--13, 2026, Jeju Island, Republic of Korea}
\acmPrice{}
\acmDOI{10.1145/3770854.3783936}
\acmISBN{979-8-4007-2258-5/2026/08}

\acmSubmissionID{v1ads0109}

\usepackage{algorithm}
\usepackage{algorithmic}
\usepackage{makecell}
\usepackage{graphics}
\usepackage{subcaption}
\usepackage{multirow}
\usepackage{float}
\usepackage{enumitem}
\usepackage{tikz}
\usepackage{balance}
\usepackage{placeins}
\newcommand{\tikzxmark}{%
\tikz[scale=0.23] {
    \draw[line width=0.7,line cap=round] (0,0) to [bend left=6] (1,1);
    \draw[line width=0.7,line cap=round] (0.2,0.95) to [bend right=3] (0.8,0.05);
}}
\newcommand{\tikzcmark}{%
\tikz[scale=0.23] {
    \draw[line width=0.7,line cap=round] (0.25,0) to [bend left=10] (1,1);
    \draw[line width=0.8,line cap=round] (0,0.35) to [bend right=1] (0.23,0);
}}
\usepackage{stfloats}

\begin{document}

\title[Dynamic Content Moderation in Livestreams]{Dynamic Content Moderation in Livestreams: Combining Supervised Classification with MLLM-Boosted Similarity Matching}

\author{Wei Chee Yew}
\email{weichee.yew@tiktok.com}
\orcid{0000-0002-1924-7412}
\affiliation{%
 \institution{TikTok}
 \city{Singapore}
 \country{Singapore}
}

\author{Hailun Xu}
\email{xuhailun@tiktok.com}
\orcid{0009-0007-9421-3697}
\affiliation{%
 \institution{TikTok}
 \city{Singapore}
 \country{Singapore}
}

\author{Sanjay Saha}
\email{sanjaysaha@tiktok.com}
\orcid{0009-0003-5386-0768}
\affiliation{%
 \institution{TikTok}
 \city{Singapore}
 \country{Singapore}
}

\author{Xiaotian Fan}
\email{xiaotian.fan@tiktok.com}
\orcid{0009-0005-4087-3353}
\affiliation{%
 \institution{TikTok}
 \city{Singapore}
 \country{Singapore}
}

\author{Hiok Hian Ong}
\email{hiok.ong@tiktok.com}
\orcid{0009-0000-8104-7385}
\affiliation{%
 \institution{TikTok}
 \city{Singapore}
 \country{Singapore}
}

\author{David Yuchen Wang}
\email{david.w@tiktok.com}
\orcid{0009-0006-6267-5291}
\affiliation{%
 \institution{TikTok}
 \city{Singapore}
 \country{Singapore}
}

\author{Kanchan Sarkar}
\authornote{Corresponding author.}
\email{kanchan.sarkar@tiktok.com}
\orcid{0009-0008-0413-2541}
\affiliation{%
 \institution{TikTok}
 \city{San Jose}
 \country{United States}
}

\author{Zhenheng Yang}
\email{zhenheny@gmail.com}
\orcid{0000-0003-0303-5885}
\affiliation{%
 \institution{TikTok}
 \city{San Jose}
 \country{United States}
}

\author{Danhui Guan}
\email{guandanhui@bytedance.com}
\orcid{0009-0009-3540-8608}
\affiliation{%
 \institution{TikTok}
 \city{Shanghai}
 \country{China}
}


\renewcommand{\shortauthors}{Wei Chee Yew et al.}
\begin{abstract}
  Content moderation remains a critical yet challenging task for large-scale user-generated video platforms, especially in livestreaming environments where moderation must be timely, multimodal, and robust to evolving forms of unwanted content. We present a hybrid moderation framework deployed at production scale that combines supervised classification for known violations with reference-based similarity matching for novel or subtle cases. This hybrid design enables robust detection of both explicit violations and novel edge cases that evade traditional classifiers. Multimodal inputs (text, audio, visual) are processed through both pipelines, with a multimodal large language model (MLLM) distilling knowledge into each to boost accuracy while keeping inference lightweight. In production, the classification pipeline achieves $67\%$ recall at $80\%$ precision, and the similarity pipeline achieves $76\%$ recall at $80\%$ precision. Large-scale A/B tests show a \textbf{6–8\% reduction in user views of unwanted livestreams}. These results demonstrate a scalable and adaptable approach to multimodal content governance, capable of addressing both explicit violations and emerging adversarial behaviors.

\end{abstract}





\maketitle

\section{Introduction}
Video streaming platforms have experienced exponential growth in recent years\cite{tiktok2024live, statista2025livestreaming}, becoming a dominant form of digital media consumption. While fixed-length videos remain popular, live streaming has emerged as an equally, if not more compelling feature, driven primarily by live e-commerce, video game streaming, and real-time interactive content. However, as live streaming gains traction, ensuring high-quality content delivery while preventing harmful or policy-violating material from being recommended or broadcast has become a critical challenge.

Although content moderation for pre-recorded short videos has been well studied in literature, research focusing specifically on live-streaming moderation remains scarce. The real-time, dynamic, and highly interactive nature of live streaming introduces unique technical and operational challenges that differ significantly from those of static video platforms. Livestreams are also much longer in duration, significantly increasing computational demands and requiring continuous, real-time monitoring. These constraints make it far more difficult to deploy large models directly in the online pipeline, necessitating lighter, optimized approaches. This paper seeks to address this gap by exploring effective content moderation strategies tailored to the demands of live-streaming platforms.

In this work, we propose a hybrid moderation framework that integrates two complementary paradigms: preset violation detection based on supervised multiclass classification and reference-based similarity matching based on video-clip retrieval and re-ranking. Our architecture is designed to address the dual need for precision in known violation categories and flexibility in identifying emerging or ambiguous edge cases. The system ingests multimodal live-stream inputs, processing them through a dual-path architecture. The first path employs a supervised classification model trained on labeled examples to detect explicit, high-confidence violations. The second path leverages a similarity matching engine that compares incoming content to a curated set of known policy violations using semantic and perceptual embeddings, enabling the system to generalize to previously unseen behaviors through nearest-neighbor retrieval. The two paths are complimentary as there is a large portion of detections are exclusive. Specifically, reference matching contributes approximately 22\% additional coverage beyond the classification branch. While our offline benchmarks show that a finetuned multimodal large language–vision model (MLLM) would offer the strongest performance, it is not feasible to deploy such a model in production due to resource constraints. Instead, we adopt a knowledge distillation approach, using the MLLM as a teacher to guide and enhance the performance of preset violation detection model and the re-ranking model during training.

We evaluate our system on a production-scale live streaming platform, operating in a real-world environment with millions of hours of content. Empirical results show that the hybrid framework significantly improves both coverage and precision, outperforming standalone classifiers or similarity systems by a wide margin. Our contributions can be summarized as follows:

\begin{itemize}[itemsep=1pt, topsep=1pt, leftmargin=*]
    \item \textbf{Hybrid Multimodal Moderation Architecture}: We propose a novel dual-path system that unifies supervised multiclass classification with reference-based similarity retrieval, applied over text, audio, and visual modalities. This hybrid design enables high-precision detection of known violations while supporting flexible generalization to emerging and ambiguous policy breaches.
    \item \textbf{Efficient Knowledge Distillation and Contrastive\\ Pre‑training for Real-Time Use}: We leverage a distillation framework where a strong, large language–vision model (LLM) serves as a teacher to train a lightweight re-ranking model that is feasible for real-time deployment. Additionally, we employ MoCo-style contrastive pretraining with a memory bank and momentum encoder, using CLIP loss to align semantic and perceptual embeddings across modalities.
    \item \textbf{Production-Scale Deployment with Comprehensive Impact Evaluation}: We deploy our system on a large-scale live-streaming platform processing millions of hours of content. Our empirical evaluation demonstrates substantial gains in coverage at high precision, and large scale online A/B experiments show a reduction of unwanted livestream views by $6\%$ to $8\%$, demonstrating measurable impact in real-world production.
\end{itemize}
Together, these results highlight the promise of hybrid, multimodal approaches for scalable and effective moderation in fast-evolving live-streaming ecosystems.

\section{Related Works}
\subsection{Content Moderation for Live Streaming}
The prevalence of online video platforms have shifted the dynamics of business, entertainment, and information. Content moderation plays an important role to ensure quality and safety of online video platforms. Human moderation of such systems can incur significant costs, cause negative mental impact on human moderators \cite{content_moderation_book, moderator_mental_health}, and be prone to inconsistencies and biases \cite{human_bias_in_moderation}. Machine-moderation systems arise as a solution for effective content moderation. Existing methods have applied neural networks towards such tasks, such as utilizing transformers or embedding models for hate speech detection \cite{embeddings_for_hate_speech, transformer_for_hate_speech}, CNNs for detecting violative videos \cite{multi_modal_cnn_video_censorship}, and localization modules for video copy detection \cite{spd, transvcl, rtr}. 

Live-streaming has become a popluar method for content creators to engage with audiences and platforms such as TikTok, Youtube, Twitch, and Red-Note have popularized live-streaming content. Live streaming platforms pose unique challenges for content moderation due to the real-time, dynamic, and multimodal nature of the content being broadcast. Unlike pre-recorded videos or static content, live streams require immediate detection and intervention to prevent harmful or policy-violating material from reaching viewers. 

\subsection{Multimodal Learning for Video Understanding}
Recent works have applied large language models (LLMs) and have shown they can be effective at content moderation for language-related tasks \cite{policy_as_prompt, google_llm_content_moderation, llm_for_rule_based_moderation, zhu2025focus}. However, for online platforms, the challenge arises from the complexity of the media, which involve complex multi-modal signals including visual, auditory, and speech. Multi-modal LLMs (MLLMs) have explored techniques to integrate additional modalities, such as video or audio, from prevailing proprietary models such as GPT4o \cite{gpt4o} and Gemini \cite{gemini}, to recent open-source models such as InternVL \cite{internvl}, QwenAudio \cite{qwen_audio}, and Llava-One-Vision \cite{llava_ov}. As such, MLLMs have began to be applied in the context of content moderation, by integrating image/video information \cite{vlm_as_policy, multi_modal_llm_moderation}, as well as audio \cite{qarm}. The key challenge lies in  effectively utilizing the MLLM to real-world use cases in a resource-efficient way.

\subsection{Leveraging LLMs for Knowledge Distillation}
Although multimodal large language models (MLLMs) deliver superior policy-understanding and cross-modal reasoning capabilities, their size and latency prohibit direct deployment in real-time moderation systems. Knowledge distillation\cite{hinton2015distilling} (KD) offers a principled solution by transferring the rich, multimodal knowledge of a teacher LLM into a compact student model optimized for low-latency inference \cite{PengZhang2024, XuLiTao2024}. Beyond logits, secondary KD techniques—such as aligning intermediate hidden representations or relational structures—have proven useful for preserving internal model semantics \cite{SunGanCheng2020}. For instance, contrastive distillation approaches like CoDIR train the student to discriminate teacher-crafted embeddings from negative examples, improving compression efficacy \cite{SunGanCheng2020}. These methods support knowledge transfer even when student architectures differ considerably from the teacher.

In application to content moderation, distilling from a frozen LLaVA-One-Vision\cite{llava_ov} teacher enables a lightweight student re-ranking model to emulate multimodal decision boundaries learned by the full MLLM—providing strong semantic alignment and inference efficiency needed for live moderation tasks.

\begin{figure*}
    \centering
    \includegraphics[width=0.95\textwidth]{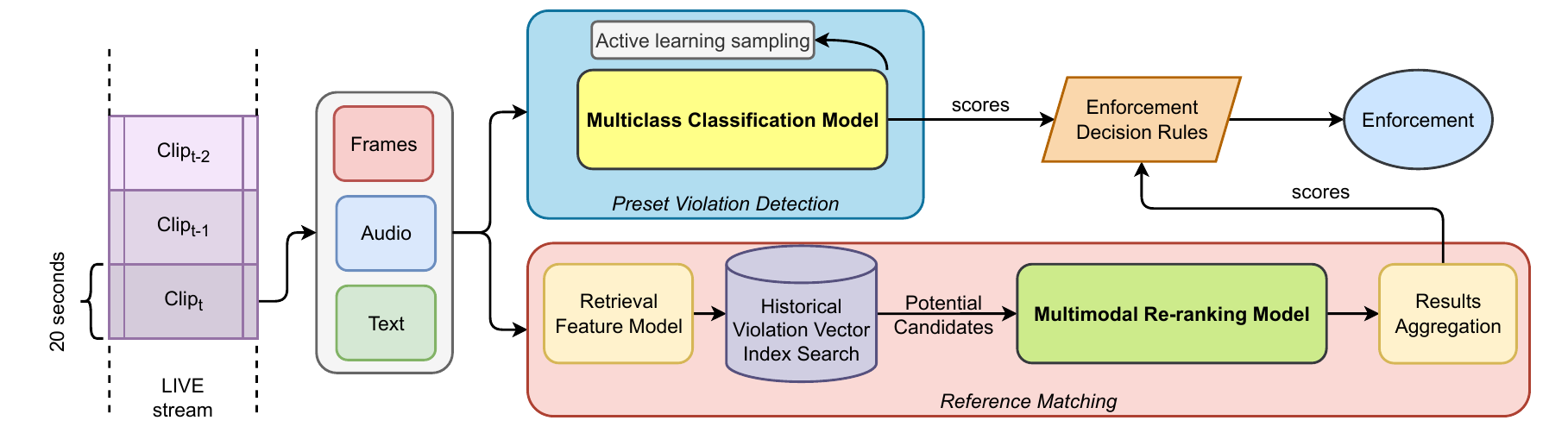}
    \caption{Overview of our proposed hybrid moderation framework. Incoming live-stream clips are segmented into 20-second windows and processed into multimodal inputs (frames, audio, text). The system follows two parallel paths: (1) Preset Violation Detection using a supervised multiclass classification model trained on labeled data, and (2) Reference Matching through a similarity-based retrieval system. Retrieved candidates are refined by a multimodal re-ranking model that evaluates semantic alignment across modalities. Enforcement decisions for Live Streams are made of predefined decision rules.}
    \label{fig:overall_framewrok}
\end{figure*}

\section{Method}
Live streaming platforms pose unique challenges for content moderation due to the continuous and transient nature of their data. To enable real-time moderation, our system processes live streams by segmenting them into manageable temporal units. Specifically, each stream is split into 20-second clips, which we refer to as query clips. The goal is to analyze each query clip independently and eventually determine whether the live stream contains any content that violates platform policies.

\subsection{Overall Framework}
Our content moderation framework follows a dual-path architecture, consisting of:
\begin{enumerate}[noitemsep, topsep=0pt]
 
    \item Preset Violation Detection Pipeline: Designed to detect known categories of policy-violating content, such as official contents or paid contents (such as major sports events, TV shows, films, music videos etc.) broadcast on non-official livestreams.
    \item Reference-based Similarity Matching Pipeline: Complements the classifier by capturing emerging or subtle patterns of violations. This system computes feature embeddings for each query clip and retrieves semantically or perceptually similar content from pre-indexed historical clips known to have violated policies.
\end{enumerate}

These two pipelines work in parallel to maximize detection of both preset copyright violations (via supervised classification) and coverage of recent violation trends (via retrieval-based matching), allowing the system to respond to policy violations in real time.

\subsection{Preset Violation Detection  Pipeline}

\begin{figure}
    \centering
    \includegraphics[width=0.5\textwidth]{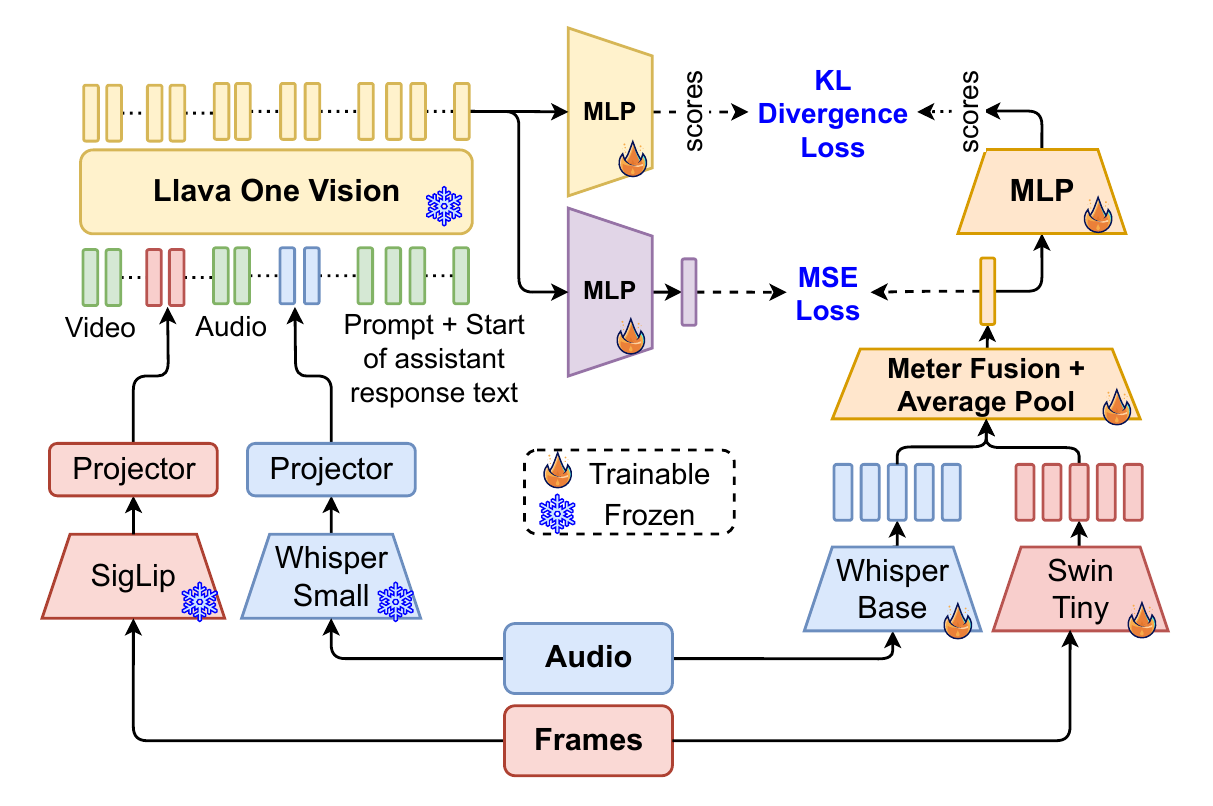}
    \caption{Architecture of the supervised classification pipeline for preset violation detection. The framework consists of a frozen LLaVA-One-Vision\cite{llava_ov} model after supervised fine-tuning (SFT) serving as a teacher model. The student model learns to align with the teacher outputs via MSE loss for last hidden state and KL Divergence loss for logits.}
    \label{fig:rc_finegrain}
\end{figure}

\textbf{Small Model Structure.}
The online Preset Violation Detection system (Figure~\ref{fig:rc_finegrain}) is powered by a lightweight multimodal model optimized for real-time inference. Visual features are extracted using a Swin-Tiny\cite{liu2021swin} visual encoder, while audio features are obtained using a Whisper-Base\cite{radford2022robust} audio encoder, followed by adaptive average pooling to reduce the number of feature embeddings. These visual and audio features are then passed to a METER\cite{dou2022meter} fusion module for multimodal integration. The fused representation is pooled and fed into a multilayer perceptron (MLP) to produce the final prediction score. For supervised training on labeled data, cross-entropy (CE) loss, $\mathcal{L}_{\text{CE}}$ is applied to the logits produced by the MLP. During knowledge distillation with unlabeled data, the model is trained using a combination of Kullback–Leibler (KL) divergence loss, $\mathcal{L}_{\text{KL}}$ and mean squared error (MSE) loss, $\mathcal{L}_{\text{MSE}}$.

\begin{equation*}
    \mathcal{L}_{\text{CE}} = - \sum_{i=1}^{C} y_i \log(p_i)
\end{equation*}
Where, $C$ is the number of classes, $y_i$ is the ground truth, $p_i$ is the predicted probability for class $i$.

\begin{equation*}
    \mathcal{L}_{\text{KL}} = \sum_{i=1}^{C} p_i \log\left( \frac{p_i}{q_i} \right)
\end{equation*}
Where, $p_i$ is the true distribution, $q_i$ is the predicted distribution, $C$ is the number of classes.
\begin{equation*}
    \mathcal{L}_{\text{MSE}} = \frac{1}{n} \sum_{i=1}^{n} (y_i - \hat{y}_i)^2
\end{equation*}
Where, $n$ is the number of data points, $y_i$ is the true value, $\hat{y}_i$ is the predicted value.

\textbf{MLLM Model Structure.}
To enable improved performance and effective transfer learning, we adopt a modified version of LLaVA-One-Vision as the teacher model. While the original architecture supports vision-text inputs, we extend it to also accommodate audio embeddings, enabling true multimodal instruction. Specifically, we modify the model’s input pipeline to integrate audio features alongside visual features. Supervised fine-tuning (SFT) is then performed using our proprietary training dataset. During SFT, the SigLip visual encoder and Whisper-Small audio encoder are kept frozen, while the language model is fine-tuned using Low-Rank Adaptation\cite{hu2021lora} (LoRA) with a standard cross-entropy loss as the training objective.

To efficiently train the smaller model, we employ a dual-objective distillation strategy from the teacher MLLM, combining hidden state distillation and logit distillation. The hidden state distillation uses Mean Squared Error (MSE) loss between the final token hidden states of the teacher and student models, while the logit distillation applies Kullback-Leibler (KL) divergence between their predicted logits. This combined approach enables the student model to maintain the teacher's generalization capabilities while meeting strict low-latency requirements, ensuring both performance and efficiency in deployment.

\textbf{Active Learning.}
To continuously enhance model performance, we have deployed an active learning module based on the Info-Coevolution\cite{qin2025infoCoevolution} framework introduced in recent research. This method enables the model to evolve alongside the data by actively identifying and sampling informative, edge-case, or difficult instances from production streams. Specifically, we use information loss as a selection criterion to prioritize samples that are expected to provide the highest learning value upon annotation. These high-value samples are then sent for human annotation and added to the training set, facilitating targeted model refinement and improved generalization over time.


\begin{figure}
    \centering
    \includegraphics[width=0.5\textwidth]{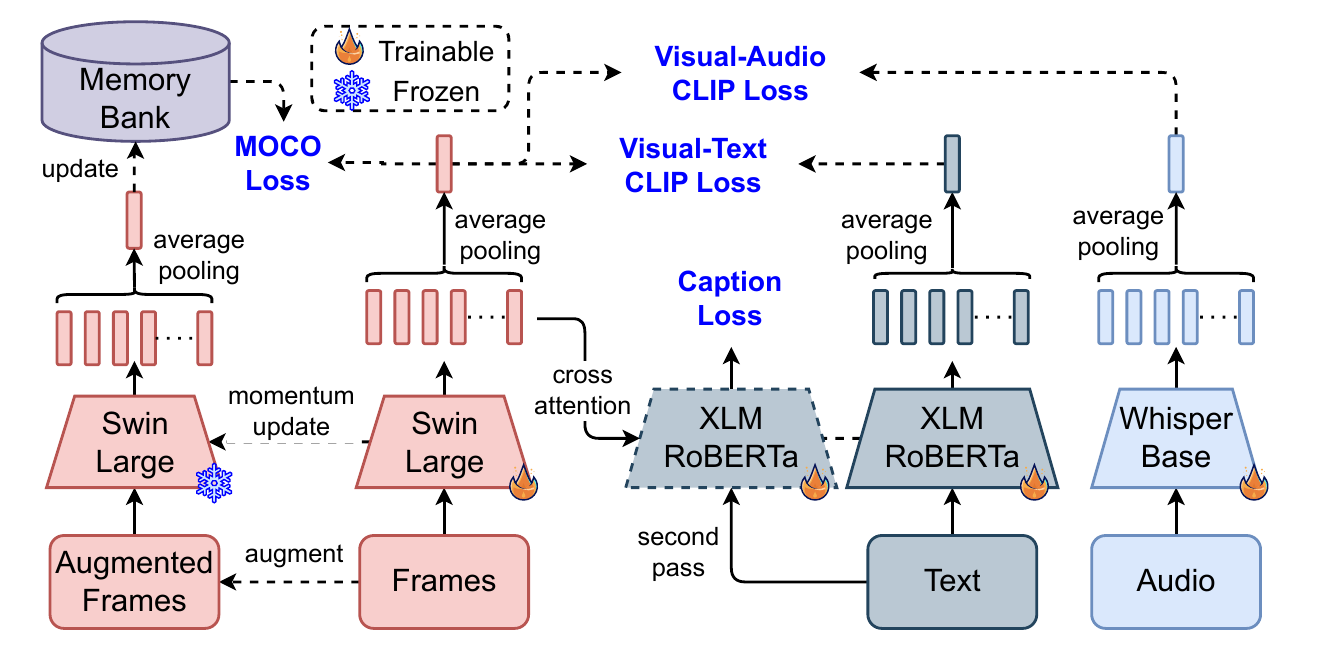}
    \caption{Training pipeline for the video-clip retrieval feature model. The visual encoder is trained using the MoCo\cite{he2020moco} framework with a momentum encoder and memory bank. To enhance semantic richness, we incorporate CLIP\cite{radford2021clip} losses between visual-text and visual-audio embeddings for cross-modality alignment. Additionally, caption supervision is introduced via a second-pass of text decoder using cross-attention on visual features.}
    \label{fig:swinlarge_training}
\end{figure}

\begin{figure}
    \centering
    \includegraphics[width=0.5\textwidth]{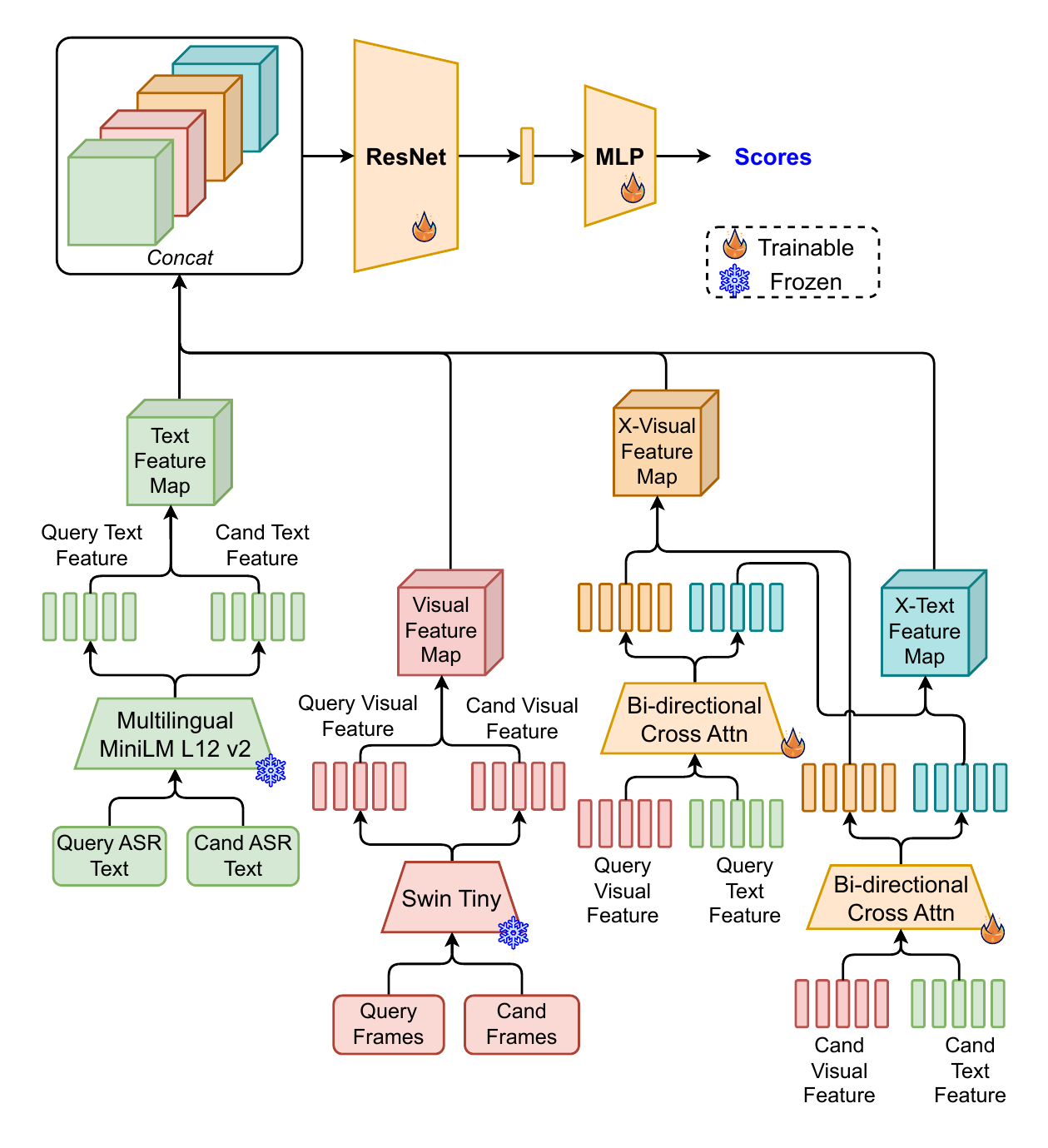}
    \caption{Small Re-ranking Model: the small/lightweight multimodal re-ranking model that integrates multimodal embeddings via bi-directional cross-attention layers. Feature maps obtained are then processed by a ResNet\cite{he2016deep}-MLP architecture for scoring. This model is also knowledge-distilled from a fine-tuned LLaVA-One-Vision model, similar to the approach shown in Figure~\ref{fig:rc_finegrain}.}
    \label{fig:matcher_architecture}
\end{figure}

\subsection{Reference-based Similarity Matching Pipeline}
To detect novel, adversarial, or edge-case violations that may not yet be covered by standard classifiers, our system employs a reference-based similarity matching pipeline. This approach ensures robust detection of policy-violating content by comparing incoming clips against a curated database of known violations.

\textbf{Video-Clip Retrieval Feature Model.}
The visual feature for candidate recall serves as a critical component in our video-clip analysis pipeline due to its efficiency and reliability in content representation. Extracted using a Swin-Large\cite{liu2021swin} model trained on a vast dataset of 4.5 billion images, combining open-source and proprietary data, these features capture rich visual semantics while maintaining computational tractability.

We trained our visual retrieval embeddings using the MoCo\cite{he2020moco} (Momentum Contrast) framework, which enables effective representation learning through contrastive learning with a dynamic memory bank by optimizing the multi-similarity loss function.

\begin{equation*}
\begin{aligned}
\mathcal{L}_{MOCO} = &\frac{1}{|\mathcal{P}|} \sum_{(i,j) \in \mathcal{P}} \log\left(1 + \sum_{(i,k) \in \mathcal{N}} e^{\alpha (s_{ik} - \lambda)}\right) \\
&+ \frac{1}{|\mathcal{N}|} \sum_{(i,k) \in \mathcal{N}} \log\left(1 + \sum_{(i,j) \in \mathcal{P}} e^{\beta (\lambda - s_{ij})}\right)
\end{aligned}
\end{equation*}

\[
\text{where:}
\begin{cases}
\mathcal{P} = \{(i,j) \mid \text{$i$ and $j$ form a positive pair}\} \\
\mathcal{N} = \{(i,k) \mid \text{$i$ and $k$ form a negative pair}\} \\
s_{ij} = \text{cosine similarity between samples $i$ and $j$} \\
\alpha = \text{weighting hyperparameters for negatives} \\
\beta = \text{weighting hyperparameters for positives} \\
\lambda = \text{similarity margin threshold}
\end{cases}
\]

To enhance the semantic richness of these visual embeddings, we further incorporated CLIP\cite{radford2021clip} training using a learnable NT-Xent\cite{chen2020simclr} (Normalized Temperature-scaled Cross Entropy) loss, aligning visual features for video clips with their corresponding textual descriptions and audio information. This multimodal training approach allows the model to capture both visual similarity and semantic meaning, resulting in improved generalization and more accurate similarity retrieval.

\begin{equation*}
    \mathcal{L}_{\text{CLIP}} = -\frac{1}{N} \sum_{i=1}^N \left[ \log \frac{e^{\left(v_i \cdot w_i / \tau\right)}}{\sum_{j=1}^N e^{\left(v_i \cdot w_j / \tau\right)}} \;+\; \log \frac{e^{\left(w_i \cdot v_i / \tau\right)}}{\sum_{j=1}^N e^{\left(w_i \cdot v_j / \tau\right)}} \right]
\end{equation*}

Where, $v_i$ is the embedding of the $i$-th visual input, $w_i$ is the embedding of the corresponding $i$-th text or audio input depending on the modality, $\tau>0$ is a temperature scaling hyperparameter, $N$ is the batch size (number of pairs).

By combining MoCo's instance discrimination with CLIP's cross-modal alignment, the learned embeddings are not only robust for visual retrieval but also semantically aware, bridging the gap between image content and textual concepts.

To optimize flexibility and efficiency, we leverage Matryoshka Representation Learning\cite{kusupati2022matryoshka} (MRL), which enables a single model to generate embeddings of varying dimensions (e.g., 32, 64, 128, 512, 768, etc). This capability allows for dynamic resource allocation, lower-dimensional embeddings can be used for storage efficiency, while higher-dimensional embeddings are reserved for tasks demanding greater precision. This eliminates the need to train and maintain separate models for different resolution requirements.

\textbf{Vector Index.}
For fast and scalable similarity searches, we utilize Hierarchical Navigable Small World\cite{malkov2020hnsw} (HNSW) indices, which are particularly well-suited for high-dimensional vector spaces. HNSW provides low-latency query performance while maintaining high recall rates, making it ideal for real-time moderation applications. The system maintains separate HNSW indices for different violation categories, such as copyright infringement, duplicate content detection, and other policy-specific issues. This modular approach ensures that searches remain efficient and relevant to the specific moderation task at hand.

\textbf{Re-ranking model.}
The re-ranking model (Figrue~\ref{fig:matcher_architecture}) refines query - candidate clip-pairs generated during the initial retrieval stage. Although visual similarity offers a strong baseline for identifying relevant content, it frequently leads to false positives in the context of livestreams. This is primarily due to the repetitive and visually similar elements common across different streams—such as shared backgrounds, user interfaces, or recurring patterns in host appearance and behavior. These similarities can mislead a purely visual retrieval system, necessitating a more nuanced re-ranking approach that can better distinguish truly matching content from visually similar but unrelated clips.

To overcome this, we adopt a multimodal re-ranking model that combines visual features and transcribed speech (ASR) to generate a fused similarity score. This multimodal approach enables more context-aware comparisons and significantly improves ranking accuracy in borderline or ambiguous cases by reducing over-reliance on visual features alone.

To further enhance learning, we apply knowledge distillation (Figrue~\ref{fig:knowledge_distillation_mm} in Appendix) using a fine-tuned large language-vision model, the LLaVa-One-Vision\cite{li2024llava_onevision}, which provides richer semantic supervision. The student model (multimodal re-ranking model) is trained to align with the teacher's outputs, enabling it to better capture nuanced cross-modal relationships.

The overall training objective includes Cross-Entropy Loss ($\mathcal{L}_{\text{CE}}$) for direct supervision, KL Divergence Loss ($\mathcal{L}_{\text{KL}}$) over predicted logits for distribution alignment, and Mean Squared Error Loss ($\mathcal{L}_{\text{MSE}}$) on hidden states to promote representation consistency.

This setup enables robust re-ranking that integrates both semantic and perceptual signals to reduce false positives and improve moderation quality.


\textbf{Aggregation Algorithm}
Matching a single query clip to a reference clip can be noisy and error-prone, especially in live-streaming environments where short clips may not fully capture the context of a policy violation. To improve robustness and reduce false positives, we employ a clip-match aggregation algorithm (Algorithm~\ref{algo:aggregation} in Appendix) that leverages temporal continuity across multiple query-reference matches. The core idea is to identify sequence-level matches between a live room and a known violating room, based on the alignment of timestamps across matched clip pairs. By grouping matched clips and evaluating their temporal alignment within a tolerance window $\epsilon$, we can infer stronger matches at the session level rather than clip level. This strategy significantly improves accuracy by enforcing temporal consistency among detected matches.

Based on the classification scores and similarity match results, appropriate policy enforcement actions are applied. 
\vspace{-0.5em}
\section{Experiments and Results}

\subsection{Datasets}

Our models were trained on a comprehensive dataset combining large-scale in-house annotated livestream content and publicly available open-source data. The in-house data comprises extensive human annotations of policy-violating clips, curated via both random sampling and high-traffic livestream streams to capture diverse and real-world violations. Complementing this, open-source datasets, primarily LAION-2B\cite{schuhmann2022laion5b} (both English-only and multilingual subsets), provide a rich source of image-text pairs to support robust generalization in our retrieval models.

\begin{itemize}[leftmargin=*]
    \item \textbf{Preset Violation Classification model:} 2 million human - annotated clips from in-house livestream data, with preset violations as labels.
    \item \textbf{Reference Matching:}
    \begin{itemize}[leftmargin=*]
        \item \emph{Video Retrieval model:} 24 million in-house videos and livestreams clips consisting of video frames, audio and texts including video or livestream titles, video captions and OCR texts. 4.6 billion open-source image-text pairs from LAION-2B\cite{schuhmann2022laion5b} and multilingual variants.
        \item \emph{Re-ranking model:} 0.5 million human annotated in-house \\livestream clip pairs, enabling fine-grained verification of retrieved candidates.
    \end{itemize}
\end{itemize}

This rich multi-modal and multi-source data foundation helps ensure our models maintain robustness across highly dynamic and diverse livestream scenarios.

\subsection{Implementation Details}

Livestream videos were segmented into fixed-length 20-second clips to standardize input units for model processing. From each clip, we extracted and synchronized three complementary modalities: audio waveform, automatic speech recognition (ASR) transcripts, and key representative visual frames. These multimodal inputs harness complementary signals vital for effective moderation.

Training utilized NVIDIA H100 GPUs, enabling efficient handling of the massive datasets and supporting computationally intensive models, including knowledge distillation from large-scale language and vision models. To prepare the models for real-time deployment, we applied model tracing, post-training quantization (PTQ) and quantization aware training (QAT), optimizing inference speed and reducing memory and computational footprint with minimal accuracy sacrifice.

\subsection{Preset Violation Detection}

Table~\ref{tab:rc_finegrain_main} summarizes the performance of our supervised preset violation detection model. The model attains a strong Average Precision (AP) score of 75.84\% and an F1 score of 73.61\%, reflecting its balanced precision and recall. It is important for the online pipeline to have high precision, while we try to achieve higher recall in gradual iterations of the model. This strong performance demonstrates that direct supervised classification remains a highly effective strategy for known policy violations, providing rapid and reliable detection.

\begin{table}[h]
\centering
\resizebox{\columnwidth}{!}{
\begin{tabular}{ccccccc}
\toprule
&              & \multicolumn{5}{c}{\textbf{Recall @}}\\
\cmidrule(lr){3-7}
\textbf{AP} & \textbf{F1} & \textbf{P70} & \textbf{P75} & \textbf{P80} & \textbf{P85} & \textbf{P90} \\ \hline
75.84\%     & 73.61\%     & 75.63\%      & 71.40\%      & 66.70\%      & 62.93\%      & 47.37\%      \\ \bottomrule
\end{tabular}
}
\caption{Performance (AP, F1, and Recall at different precision levels) of the preset violation detection model.}
\label{tab:rc_finegrain_main}
\end{table}

\subsection{Candidate Recall for Reference Matching}

Effective candidate retrieval is crucial to guarantee that potential violations are included for detailed analysis in the later ranking stage. Table~\ref{tab:swinlarge_recall} reports recall rates of the video retrieval model at various ``Top k'' values.

The model achieves exceptionally high Recall@Top-5 (90.53\%) and Recall@Top-100 (98.99\%) for retrieving at least one relevant match, indicating that the system confidently surfaces appropriate candidate clips from billions of video segments. This search breadth ensures the reference matching path effectively captures diverse forms of violations that might be semantically or visually nuanced, even if unseen during training. Overall, the retrieval model's high recall supports strong downstream verification performance.

\begin{table}[h]
\centering
\begin{tabular}{llllll}
\toprule
\textbf{Top k }              & \textbf{5} & \textbf{10} & \textbf{20} & \textbf{50} & \textbf{100} \\ \midrule
\textbf{Recall one} & 90.53\%    & 94.11\%     & 96.38\%     & 98.37\%     & 98.99\%      \\
\textbf{Recall all} & 53.81\%    & 53.96\%     & 58.26\%     & 68.57\%     & 74.56\%      \\ \bottomrule
\end{tabular}
\caption{Recall performance (Recall One and Recall All) of the video retrieval model at different Top-K values.}
\label{tab:swinlarge_recall}
\vspace{-2.0em}

\end{table}

\subsection{Pair-level Re-ranking}

Table~\ref{tab:mmv3_main} presents results of the re-ranking stage, which refines the initial candidates retrieved by the visual similarity model by integrating multimodal features. The re-ranking model achieves an AP of 74.82\% and F1 of 73.49\%, demonstrating strong performance in filtering out false positives while promoting semantically aligned matches.

This re-ranking step is critical in production environments where high recall alone is insufficient, initial retrieval may surface visually or textually similar content that is contextually irrelevant or benign. By leveraging a fusion of audio, visual, and text features, the re-ranking model enhances precision by verifying the semantic and policy-relevant correspondence between retrieved clips and reference violations. The strong results validate the importance of this multimodal re-ranking model, ensuring that downstream moderation decisions are based on accurate, policy-grounded matches rather than superficial similarity.

\begin{table}[h]
\centering
\resizebox{\columnwidth}{!}{
\begin{tabular}{ccccccc}
\toprule
&              & \multicolumn{5}{c}{\textbf{Recall @}}\\
\cmidrule(lr){3-7}
\textbf{AP} & \textbf{F1} & \textbf{P70} & \textbf{P75} & \textbf{P80} & \textbf{P85} & \textbf{P90} \\ \hline
74.82\%     & 73.49\%     & 75.40\%      & 71.51\%      & 66.13\%      & 59.84\%      & 41.42\%      \\ \bottomrule
\end{tabular}
}

\caption{Performance of the Multimodal re-ranking model that refines retrieved results using multimodal features.}
\label{tab:mmv3_main}
\vspace{-2.0em}
\end{table}

\subsection{Ablation Results}

\subsubsection{Preset Violation Detection}

\begin{table*}[h]
\begin{tabular}{lllllllll}
\toprule
\textbf{Variation} & \textbf{Feature} & \textbf{AP} & \textbf{F1} & \textbf{R@P70} & \textbf{R@P75} & \textbf{R@P80} & \textbf{R@P85} & \textbf{R@P90} \\ \hline
MLLM                     & v   & 77.35\%     & 75.06\%     & 78.38\%      & 73.46\%      & 70.25\%      & 63.27\%      & 35.70\%      \\
MLLM                     & v + a   & 80.03\%     & 76.57\%     & 80.32\%      & 77.23\%      & 70.71\%      & 63.96\%      & 55.03\%      \\
Small                    & v           & 64.70\%     & 68.35\%     & 66.59\%      & 61.44\%      & 44.51\%      & 34.55\%      & 13.62\%      \\
Small                    & v + a   & 71.05\%     & 71.11\%     & 71.51\%      & 65.45\%      & 57.21\%      & 49.31\%      & 39.70\%      \\
Small + KD               & v + a   & 75.84\%     & 73.61\%     & 75.63\%      & 71.40\%      & 66.70\%      & 62.93\%      & 47.37\%      \\ \bottomrule
\end{tabular}
\caption{Ablation study for the preset violation detection on the effect of modality integration and knowledge distillation in preset violation detection. `v' indicates visual features only, `v+a' indicates combined visual and audio features. The ``Small'' model refers to the lightweight student trained from scratch, ``MLLM'' denotes the large teacher model, and ``Small+KD'' denotes the small model distilled from the MLLM.}

\label{tab:rc_finegrain_ablation}
\end{table*}

\begin{table*}[h]
\begin{tabular}{ccccccccc}
\toprule
\textbf{\makecell{Visual\\Backbone}} & \textbf{\makecell{Audio\\Backbone}} & \textbf{\makecell{Text\\Backbone}} & \textbf{\makecell{Fusion\\Module}} & 
\textbf{\#Params} & \textbf{QPS} & \textbf{AP} & \textbf{R@P80} & \textbf{R@P90} \\ \hline
SwinL & W-S & \tikzcmark & 12-L & 650M & 14 & 77.53\% & 68.31\% & 52.17\% \\
SwinL & W-B & \tikzcmark & 4-L & 520M & 28 & 74.90\% & 68.08\% & 41.30\% \\
SwinL & W-B & \tikzcmark & 4-L + AMP & 520M & 46 & 75.08\% & 65.56\% & 50.46\% \\
SwinL & W-B & \tikzxmark & 4-L + AMP & 240M & 56 & 73.46\% & 66.13\% & 35.47\% \\
SwinT & W-B & \tikzxmark & 4-L + AMP & 75M & 87 & 71.05\% & 57.21\% & 39.70\% \\
\bottomrule
\end{tabular}
\caption{Ablation study on the effect of model size and backbone configuration on preset violation detection performance (without knowledge distillation). The \textbf{Visual Backbone} uses either SwinL (Large) or SwinT (Tiny). The \textbf{Audio Backbone} employs Whisper-Small (W-S) or Whisper-Base (W-B). The \textbf{Text Backbone} is XLM-Roberta or omitted. Fusion modules are denoted as 12-L (12 layers) or 4-L (4 layers) with AMP indicating adaptive mean pooling on input tokens for computational efficiency. Results show trade-offs between parameter count, throughput (QPS), and detection performance (AP, R@P80, R@P90).}
\label{tab:rc_finegrain_ablation_small}
\end{table*}

\begin{table*}[h]
\centering
\begin{tabular}{cc|ccccc|ccccc}
\toprule
\multirow{2}{*}{\textbf{Feature}} & \multirow{2}{*}{\textbf{Dimension}} & \multicolumn{5}{c|}{\textbf{RecallOne @ Top-K}} & \multicolumn{5}{c}{\textbf{RecallAll @ Top-K}} \\ \cline{3-12} 
 &  & \textbf{5} & \textbf{10} & \textbf{20} & \textbf{50} & \textbf{100} & \textbf{5} & \textbf{10} & \textbf{20} & \textbf{50} & \textbf{100} \\ \hline
MoCo + CLIP & 768 & 90.72\% & 94.32\% & 96.66\% & 98.57\% & 99.18\% & 54.26\% & 54.38\% & 58.94\% & 69.26\% & 75.13\% \\
MoCo + CLIP & 512 & 90.77\% & 94.26\% & 96.60\% & 98.49\% & 99.17\% & 54.22\% & 54.29\% & 58.79\% & 69.12\% & 75.08\% \\
MoCo + CLIP & 256 & 90.66\% & 94.20\% & 96.57\% & 98.51\% & 99.15\% & 54.10\% & 54.23\% & 58.56\% & 68.83\% & 74.67\% \\
MoCo + CLIP & 128 & 90.53\% & 94.11\% & 96.38\% & 98.37\% & 98.99\% & 53.81\% & 53.96\% & 58.26\% & 68.57\% & 74.56\% \\
MoCo & 128 & 59.31\% & 66.30\% & 73.10\% & 81.25\% & 86.14\% & 32.19\% & 33.05\% & 38.42\% & 50.05\% & 58.47\% \\ \bottomrule
\end{tabular}
\caption{Recall performance of the retrieval model across varying embedding dimensions (128–768) for different feature configurations. The combination of MoCo and CLIP consistently achieves high recall across all Top-K settings, while reduced-dimensional embeddings preserve accuracy with improved efficiency, enabling scalable production deployment.}
\label{tab:swinlarge_recall_ablations}
\end{table*}

Table~\ref{tab:rc_finegrain_ablation} compares variants of the preset violation classification model to assess the impact of modality inclusion and knowledge distillation. Using only visual inputs yields lower performances on both small model (AP=64.70\%) and MLLM (AP=77.35\%). Incorporating audio modality boosts AP by 6.3\% on small model and 2.7\% on MLLM, underscoring the importance of multimodal signals in ambiguous scenarios.

The MLLM-based model attains the highest AP (80.03\%), demonstrating the power of large multimodal language models. However, its computational cost makes it impractical for real-time production. Our distilled Small+KD model strikes a strong balance, achieving 75.84\% AP by transferring knowledge from the MLLM into a resource-efficient architecture. This knowledge distillation proves critical for maintaining high accuracy under tight latency constraints.

In the ablation study focusing on small-scale preset‑violation detection models, several trends emerge from scaling the backbone sizes and applying adaptive mean pooling (AMP) as shown in Table~\ref{tab:rc_finegrain_ablation_small}. In the heaviest multimodal setup, with Swin Large visual backbone, Whisper Small audio backbone, XLM-RoBERTa text backbone and 12-layered METER fusion, the model achieves its best performance (AP=77.35\%) but with notably low throughput of only 14 QPS on a A10 GPU. Scaling down backbone sizes, trimming the fusion layers, and dropping the text branch accelerates inference substantially, yet each simplification brings a predictable dip in effectiveness. Adding adaptive mean pooling before METER fusion module increases model throughput and helps to slightly recover model performance, especially at higher precision level (R@P90 +9\%).

While leaner models offer impressive speed, their lower performances can be effectively countered via knowledge distillation. In the smallest student model setting, the AP jumped from around 71\% to 75.8\%, bringing it much closer to the performance of the largest model setting above, while maintaining much higher throughput suitable for online deployment. This demonstrates that distillation enables lean architectures to serve traffic at scale with minimal loss in effectiveness relative to the largest multimodal configuration.

\subsubsection{Candidate Recall}

Table~\ref{tab:swinlarge_recall_ablations} studies architecture and feature dimension for the retrieval model. Reducing feature dimensionality from 768 down to 128 has negligible impact on recall. This suggests that compressed embeddings effectively retain semantic information, enabling faster retrieval and ranking without sacrificing accuracy—a practical consideration for production scalability.

Introducing multi-modality CLIP constrastive loss on top of MoCo-based training framework significantly elevates recall performance. By jointly aligning image and text embeddings, the model benefits from richer semantic supervision beyond MoCo’s visual-only contrastive training. This performance gap is substantially huge especially at lower top-K (e.g. at RecallOne@Top‑5, MoCo + CLIP achieves over 90\%, whereas MoCo alone achieves only around 60\%). The gap narrows at higher top-K thresholds, but remains non-negligible.

\subsubsection{Re-ranking and Aggregation}

Table~\ref{tab:mm_matcher_ablation} evaluates the re-ranking model with and without knowledge distillation. The Small+KD model achieves a compelling AP of 74.82\%, significantly outperforming the non-distilled Small baseline (71.05\%) while remaining more efficient than the full MLLM (77.60\%). This validates the effectiveness of knowledge distillation in transferring rich multimodal semantics into lighter architectures, crucial for scalable real-time moderation.

\begin{table}[]
\resizebox{\columnwidth}{!}{%

\begin{tabular}{ccccccc}
\toprule
 &  & \textbf{} & \multicolumn{4}{c}{\textbf{Recall @}} \\ \cline{4-7} 
\textbf{Variation} & \textbf{AP} & \textbf{F1} & \textbf{P75} & \textbf{P80} & \textbf{P85} & \textbf{P90} \\ \hline
MLLM & 77.60\% & 76.54\% & 75.51\% & 72.54\% & 63.62\% & 56.86\% \\
Small & 71.05\% & 71.11\% & 65.45\% & 57.21\% & 49.31\% & 39.70\% \\
Small + KD & 74.82\% & 73.49\% & 71.51\% & 66.13\% & 59.84\% & 41.42\% \\ 
\bottomrule

\end{tabular}
}
\caption{Performance comparison of the re-ranking model across variations. The full MLLM achieves the highest accuracy but is impractical for real-time deployment due to resource constraints. The Small model offers higher efficiency but lower accuracy, while the knowledge-distilled Small+KD model significantly recovers performance, narrowing the gap to the MLLM while remaining lightweight for production.}
\label{tab:mm_matcher_ablation}
\end{table}

\subsection{Online Performance}
A key objective of our moderation pipeline is to effectively reduce user exposure to unwanted or policy-violating live streams on the platform. To evaluate real-world impact, we conducted rigorous online A/B tests following the deployment of each pipeline upgrade. These experiments measured critical business metrics related to the consumption of undesirable content, demonstrating meaningful gains over previous baselines.

The online A/B experiment was conducted over a few weeks with $10\%$ traffic assigned to each of the experiment groups. The evaluation was performed using a two-sample z-test for proportions to measure the change in view rate of violated live streams.

\begin{itemize}
    \item \textbf{Metric}: $1.2\%$ decrease in violation-related user views.
    \item \textbf{Standard deviation}: $0.055$ over the sampled population.
    \item \textbf{Confidence interval} ($95\%$): $[ -  0.1476\%, - 0.087\%]$.
    \item \textbf{P-value}: $\sim0$.
    \item \textbf{Minimum Detectable Effect} (MDE): $0.04\%$.

\end{itemize}

\textbf{Preset Violation Detection Pipeline:} The baseline in this experiment was an older moderation system that relied solely on visual features for detection. Our upgraded pipeline, which incorporates multimodal inputs (visual, audio, and ASR text) alongside knowledge distillation techniques, led to a statistically significant reduction in user views of preset violation-based live streams by 1.2\%. Additionally, there was a 2.7\% decrease in the total duration users spent watching these flagged streams. These improvements highlight how leveraging richer multimodal signals in classification enhances the precision and effectiveness of direct violation detection in a live-streaming context.
    
\textbf{Reference Matching Pipeline:}   The previous version of this pipeline employed an outdated recall model and a re-ranking module based exclusively on visual features. The updated pipeline integrates multimodal learning in both retrieval and re-ranking stages, accompanied by knowledge distillation from large multimodal language models. This upgrade yielded several notable outcomes:
\begin{itemize}[leftmargin=*]
    \item A 0.6\% reduction in user views of unwanted live streams, indicating more accurate identification and filtering of problematic content.
    \item A 0.25\% decrease in streams flagged for sensitive issues, suggesting enhanced robustness in capturing subtle or context-dependent violations.
    \item A 2.33\% reduction in duplicate live stream views relative to the baseline, underscoring improved detection and consolidation of near-duplicate or re-posted violating content—an important factor in reducing noise and user fatigue.
\end{itemize}

\textbf{Hit Latency:} 
\begin{table}[h]
\centering
\begin{tabular}{l c c c c}
\toprule
& & \multicolumn{3}{c}{\textbf{Percentile (ms)}} \\
\cmidrule(lr){3-5}
\textbf{Module} & \textbf{Average (ms)} & \textbf{50th} & \textbf{90th} & \textbf{99th} \\
\midrule
Preset Violation   & 417.55  & 350.06  & 703.42  & 1000.00 \\
Ref. Matching & 3980.00 & 4120.00 & 5490.00 & 6310.00 \\
\bottomrule
\end{tabular}
\caption{Runtime latency evaluation per 20 second clip for the two pipelines. On average, the preset violation branch's latency is about 0.4\,s, while the reference-matching branch's latency is about 4\,s,  which enables very quick detection for live streams.}
\label{tab:latency_evaluation}
\end{table}

To monitor the impact and effectiveness of our pipeline, a long-term A/B backtest experiment is carried out to monitor the performances of the preset violation detection pipeline and reference matching pipeline. The preset violation detection pipeline alone leads to about $4\%$ reduction in user views of unwanted live streams. Introducing reference matching pipeline further reduces the user views of unwanted live streams down by about $2\%$ to $4\%$, demonstrating how the enhanced reference matching pipeline could effectively complement the preset violation detection pipeline to improve overall moderation efficacy.

To minimize false positives and over-moderation, the system employs several safeguards:
\begin{itemize}
    \item Model thresholds are tuned at P90 precision to minimize overkill in automated moderation.
    \item Human moderators review detections in lower precision bands (P70-P85) for verification before enforcement.
    \item For affected creators, an appeal mechanism routes disputed cases directly to the human moderation team, where decisions are re-evaluated based on both model outputs and contextual review.
\end{itemize}

Overall, the online performance evaluation confirms that our unified dual-path moderation framework not only advances detection accuracy in offline metrics but also delivers tangible business value by reducing exposure to harmful and duplicative live streams in a production environment. The multimodal, knowledge-distilled models enable more precise and scalable content governance, thereby enhancing platform health and user experience.
\section{Conclusion}
In this work we present a unified, multimodal content moderation framework tailored for the unique demands of real-time live-streaming platforms. By integrating supervised classification with a reference-based similarity matching pipeline enhanced through knowledge distillation from large language–vision models, our approach effectively balances precision and recall to govern a rapidly evolving content ecosystem.

Our experiments demonstrate that supervised classification excels at detecting known, preset categories of policy violations with high precision and balanced recall. The inclusion of multimodal signals—visual frames, audio features, and ASR transcription—notably improved detection accuracy compared to visual-only baselines. This underscores the importance of leveraging complementary modalities to address the ambiguity and complexity often inherent in live stream content.

At the same time, the reference matching pipeline extends the system's capability to identify novel or subtle violations beyond the reach of supervised models. Leveraging a large-scale retrieval model followed by a multimodal re-ranking step guided by distillation from a large language–vision model, this pipeline achieves outstanding recall while maintaining precise verification of candidates. The ability to compress embeddings without significant recall loss further enhances retrieval efficiency and scalability, crucial for handling billions of retrospective and live clips.

The dual-path architecture offers compelling synergy: high-confidence classification enforces known policies reliably, while similarity-based retrieval generalizes to emerging violation forms and adversarial behavior. This complementarity is reflected in both offline benchmarks and online A/B testing, where pipeline upgrades led to substantial reductions in unwanted live stream views, flagged content, and duplicate stream consumption. These real-world gains translate directly into improved platform integrity and user experience.

Knowledge distillation emerges as a pivotal technique, enabling compact student models to approach the rich semantic awareness of large language–vision teachers. This transfer of multimodal contextual understanding permits efficient real-time inference, meeting tight latency constraints without sacrificing predictive power. Such scalability is imperative for production deployment in large live streaming ecosystems.

Looking ahead, this framework lays the foundation for continued innovation in multimodal moderation. Future research could explore expanding modality coverage (e.g., integrating chat signals or user interaction metadata), incorporating temporal modeling for more holistic stream-level analysis, and adapting knowledge distillation strategies to evolving LLM architectures. Additionally, enhanced active learning loops to incorporate feedback from human moderators and user reports can drive continual refinement and robustness against adversarial attempts.


\bibliographystyle{ACM-Reference-Format}
\balance
\bibliography{chapters/references}

\appendix
\section{Appendix}
\label{sec:agg_algo}

\subsection{Clip-Match Aggregation Algorithm}

Post-processing step to improve re-ranking precision by verifying prolonged matches between a query livestream and candidate reference streams. Consecutive aligned clip matches within a temporal tolerance window are aggregated, reducing false positives from isolated or spurious clip matches.
\setlength{\textfloatsep}{5pt}   
\setlength{\floatsep}{5pt}
\begin{algorithm}[h]
\footnotesize
\caption{Clip-Match Aggregation Algorithm}
\begin{algorithmic}[1]
\REQUIRE New Query clip $q_i$, Reference clips from same LIVE stream $C = \{c_1, c_2, \dots, c_m\}$, Match scores $S(q_i, c_j)$, score threshold $\tau$, temporal tolerance $\epsilon$
\ENSURE Aggregated matches between query and reference rooms

\STATE List of existing valid match pairs $M$

\STATE Current maximum match length $L_{max}$
\FOR{each candidate reference clip $c_j \in C$}
    \IF{$S(q_i, c_j) \geq \tau$}
        \STATE $L \leftarrow 1$
        \STATE $TotalScore \leftarrow S(q_i, c_j)$
        \FOR{$(q_x, c_y)$ in M}
            \IF{$|(q_i - q_x) - (c_j - c_y)| < \epsilon$}
                \STATE $L \leftarrow L+1$
                \STATE $TotalScore \leftarrow TotalScore + S(q_x, c_y)$
            \ENDIF
        \ENDFOR
        \IF{$L > L_{max}$}
            \STATE $L_{max} \leftarrow L$
            \STATE $AggScore \leftarrow TotalScore \div L$
        \ELSIF{$L = L_{max}$}
            \STATE $AggScore \leftarrow max(AggScore, TotalScore \div L)$
        \ENDIF
        \STATE Add $(q_i, c_j)$ to $M$
        
    \ENDIF
\ENDFOR
\RETURN $AggScore, L_{max}$
\end{algorithmic}
\label{algo:aggregation}
\end{algorithm}


\subsection{More Ablation Results}

\begin{table}[H]
\centering
\resizebox{\columnwidth}{!}{%
\begin{tabular}{lcccc}
\toprule
 & \multicolumn{4}{c}{\textbf{Aggregation Clip Length}} \\
\cmidrule(lr){2-5}
\textbf{Metric} & \textbf{No Aggr.} & \textbf{5} & \textbf{10} & \textbf{15} \\
\midrule
Model Hit Rate       & 0.39\% & 0.48\% & 0.41\% & 0.36\% \\
True Detection Rate  & 0.35\% & 0.44\% & 0.38\% & 0.34\% \\
Overkill Rate        & 0.04\% & 0.04\% & 0.03\% & 0.01\% \\
\bottomrule
\end{tabular}
}
\caption{Sensitivity of the system to hyperparameters (e.g., clip length, embedding dimension). Aggregation improves recall (0.35\% to 0.44\%) at similar overkill rates, allowing us to lower the similarity threshold and achieve a higher true-positive rate at similar overkill levels, while longer windows balance precision and recall by slightly reducing both true-positives and false-positives.}
\label{tab:aggregation_sensitivity}
\end{table}

\begin{table}[H]
\centering
\resizebox{\columnwidth}{!}{%
\begin{tabular}{cccccccc}
\toprule
\textbf{Aggr.} & \textbf{AP} & \textbf{F1} & \textbf{P70} & \textbf{P75} & \textbf{P80} & \textbf{P85} & \textbf{P90} \\
\midrule
No aggr.   & 75.84\% & 73.61\% & 75.63\% & 71.40\% & 66.70\% & 62.93\% & 47.37\% \\
1 min & 76.84\% & 74.51\% & 76.85\% & 71.44\% & 68.25\% & 67.69\% & 46.85\% \\
2 min & 77.68\% & 75.01\% & 77.79\% & 72.84\% & 68.96\% & 66.15\% & 46.11\% \\
3 min & 78.15\% & 75.42\% & 79.18\% & 73.20\% & 70.15\% & 69.81\% & 45.61\% \\
4 min & 78.30\% & 75.83\% & 78.84\% & 73.54\% & 70.55\% & 70.30\% & 45.91\% \\
5 min & 78.49\% & 76.06\% & 79.62\% & 73.83\% & 70.01\% & 70.55\% & 46.10\% \\
\bottomrule
\end{tabular}
}
\caption{Score aggregation of the reference matching branch: sliding windows of different durations. Aggregation boosts AP(75.8 to 78.5) and F1(73.6 to 76.1), showing temporal smoothing improves stability. In deployment, we adopt a simple sliding-window aggregation with mean pooling.}
\label{tab:aggregation_window_sensitivity}
\end{table}

\begin{table}[h]
\centering
\begin{tabular}{lcc}
\toprule
\textbf{Model} & \textbf{AP} & \textbf{F1} \\
\midrule
Gemini-2.5-Pro & 8.29\%  & 15.32\% \\
Our Approach  & 75.84\% & 73.61\% \\
\bottomrule
\end{tabular}
\caption{Comparison with a pretrained MLLM (Gemini-2.5-Pro) on the classification branch. The pretrained model exhibits extremely low precision and F1 when applied directly to the moderation domain, likely due to hallucination tendencies and a lack of task-specific grounding.}
\label{tab:gemini_classification}
\end{table}

\begin{table}[h]
\centering
\begin{tabular}{lcc}
\toprule
\textbf{Model} & \textbf{AP} & \textbf{F1} \\
\midrule
Gemini-2.5-Pro & 30.28\% & 35.84\% \\
Our Approach  & 74.82\% & 73.49\% \\
\bottomrule
\end{tabular}
\caption{Comparison with a pretrained MLLM (Gemini-2.5-Pro) on the reference-matching branch. While the pretrained model performs better than in the classification setting, it still significantly underperforms our approach, highlighting the importance of task-specific modeling and grounding for reliable moderation.}
\label{tab:gemini_reference}
\end{table}

\subsection{Exploration of Score Distribution}
\begin{figure}[H]
\centering
    \begin{subfigure}{.5\textwidth}
    \centering
    \includegraphics[width=0.95\columnwidth]{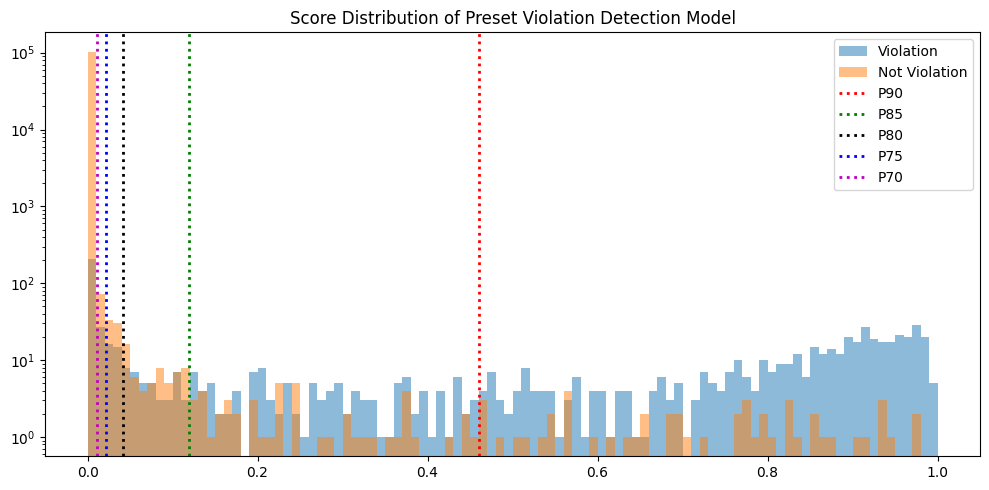}
    \caption{Preset Violation Detection model}
    \label{fig:score_distribution_preset_violation}
    \end{subfigure}%

    \begin{subfigure}{.5\textwidth}
    \centering
    \includegraphics[width=0.95\columnwidth]{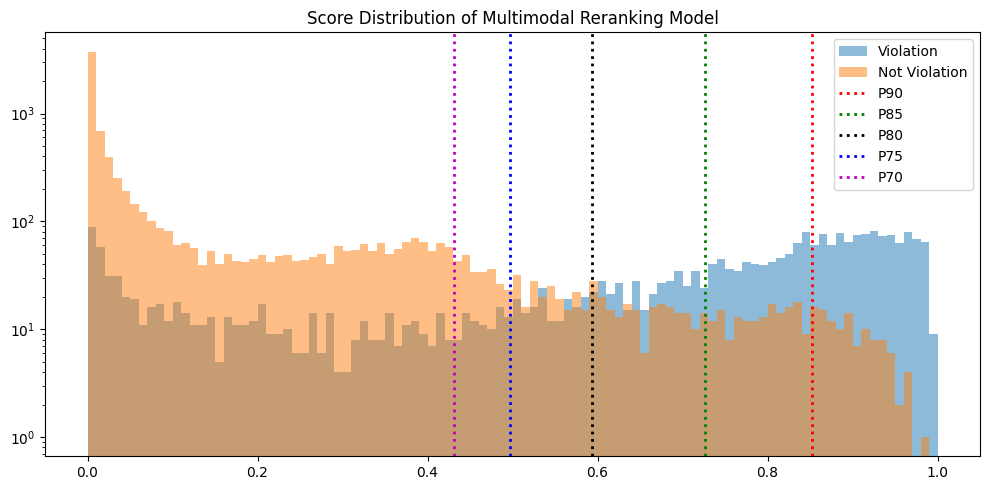}
    \caption{Reference Matching model}
    \label{fig:score_distribution_multimodal_reranking}
    \end{subfigure}
\caption{We observe that raising precision from $80\%$ to $90\%$ requires a steep threshold increase, as 'violation' and 'non-violation' scores heavily overlap in this range, indicating most hard cases lie here. And, both precision and recall remain stable around $75-80\%$, with minor fluctuations, demonstrating model stability.}
\end{figure}

\subsection{Knowledge Distillation of Re-ranking model}

\begin{figure}[H]
    \centering
    \includegraphics[width=0.5\textwidth]{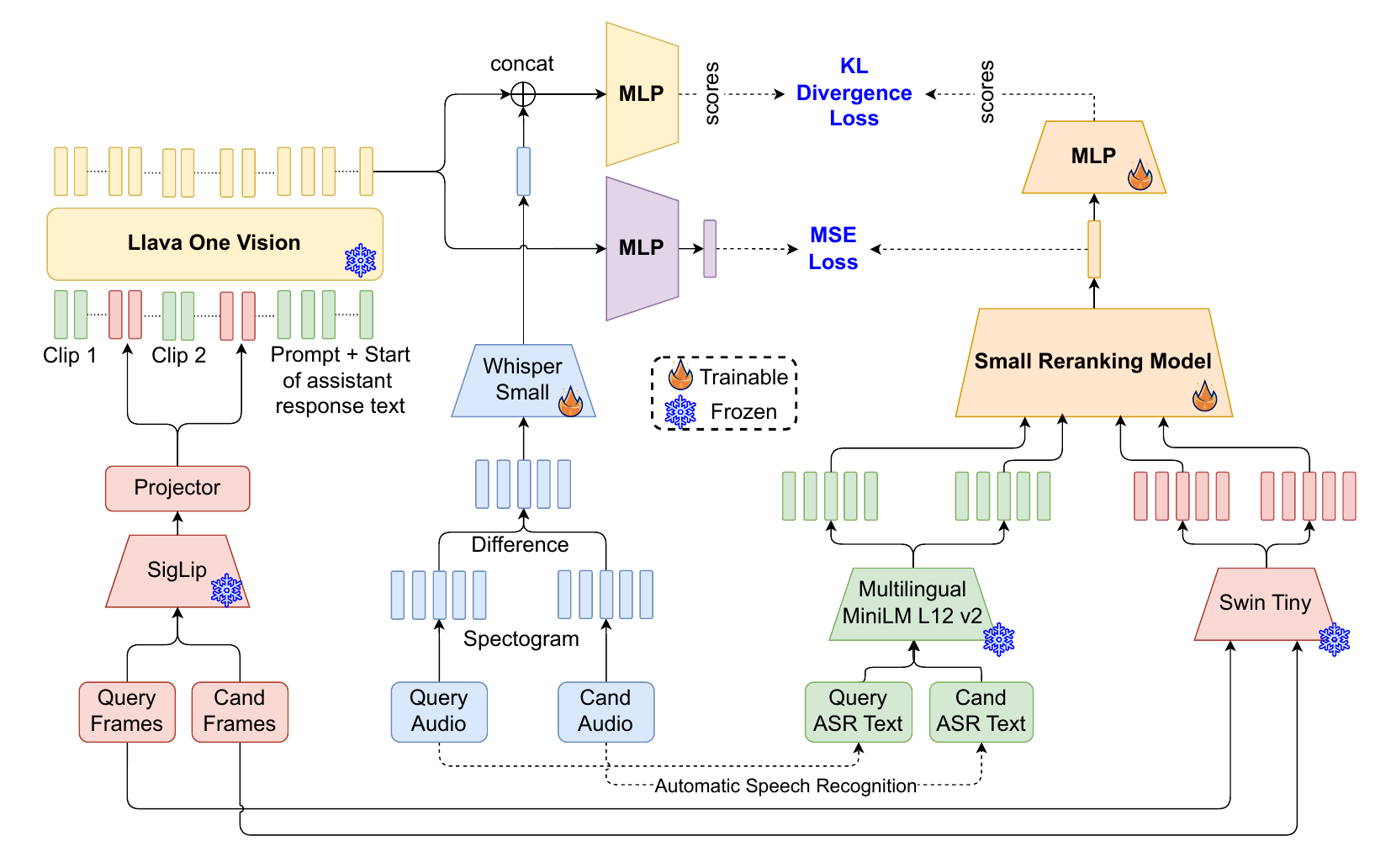}
    \caption{Knowledge Distillation of Re-ranking model: the knowledge distillation process where a frozen LLaVA-One-Vision\cite{li2024llava_onevision} teacher provides soft labels to guide the student re-ranking model. The student (Small Re-ranking Model) is trained with KL divergence and MSE losses to match teacher predictions for efficient deployment.}
    \label{fig:knowledge_distillation_mm}
\end{figure}


\end{document}